\def\year{2019}\relax
\documentclass[letterpaper]{article} 
\usepackage{aaai19}  
\usepackage{times}  
\usepackage{helvet}  
\usepackage{courier}  
\usepackage{url}  
\usepackage{graphicx}  
\usepackage{amsmath}
\usepackage{multirow}
\usepackage{amssymb}
\usepackage{array}

\title{Title-Guided Encoding for Keyphrase Generation}
\author{Wang Chen,\textsuperscript{1,2} Yifan Gao,\textsuperscript{1,2} Jiani Zhang,\textsuperscript{1,2} Irwin King,\textsuperscript{1,2} Michael R. Lyu\textsuperscript{1,2} \\
  \textsuperscript{1}Department of Computer Science and Engineering, \\
The Chinese University of Hong Kong, Shatin, N.T., Hong Kong \\
\textsuperscript{2}Shenzhen Key Laboratory of Rich Media Big Data Analytics and Application, \\
Shenzhen Research Institute, The Chinese University of Hong Kong, Shenzhen, China \\
  \{wchen, yfgao, jnzhang, king, lyu\}@cse.cuhk.edu.hk \\
  }

\begin{document}
%
\maketitle

\begin{abstract}
Keyphrase generation (KG) aims to generate a set of keyphrases given a document, which is a fundamental task in natural language processing (NLP). Most previous methods solve this problem in an extractive manner, while recently, several attempts are made under the generative setting using deep neural networks. However, the state-of-the-art generative methods simply treat the document title and the document main body equally, ignoring the leading role of the title to the overall document. To solve this problem, we introduce a new model called Title-Guided Network (TG-Net) for automatic keyphrase generation task based on the encoder-decoder architecture with two new features: (\textrm{i}) the title is additionally employed as a query-like input, and (\textrm{ii}) a title-guided encoder gathers the relevant information from the title to each word in the document. Experiments on a range of KG datasets demonstrate that our model outperforms the state-of-the-art models with a large margin, especially for documents with either very low or very high title length ratios.
\end{abstract}

\section{Introduction}
\begin{figure}
\centering
\includegraphics[width=3.2in]{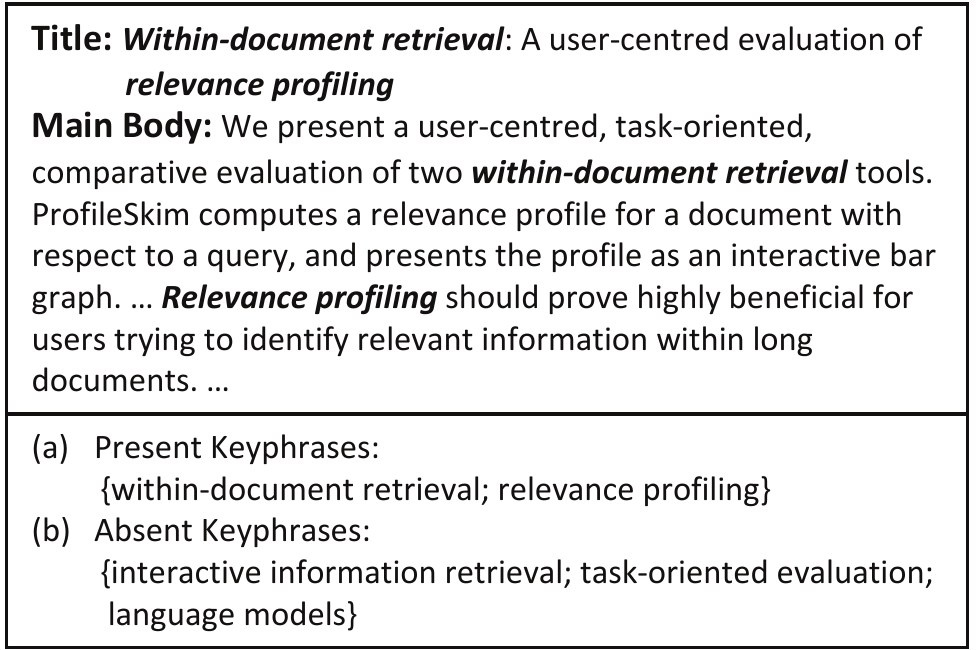}
\caption{An example of keyphrase generation. The present keyphrases are bold and italic in the source text.}
\label{figure: keyphrase generation example}
\end{figure}

Keyphrases are short phrases that can quickly provide the main information of a given document (the terms ``document'', ``source text'' and ``context'' are interchangeable in this study, and all of them represent the concatenation of the title and the main body.). Because of the succinct and accurate expression, keyphrases are widely used in information retrieval~\cite{jones1999phrasier}, document categorizing~\cite{hulth2006study}, opinion mining~\cite{berend2011opinion}, etc. Due to the huge potential value, various automatic keyphrase extraction and generation methods have been developed. As shown in Figure~\ref{figure: keyphrase generation example}, the input usually consists of the title and the main body, and the output is a set of keyphrases. 

Most typical automatic keyphrase extraction methods~\cite{witten1999kea,medelyan2009human,mihalcea2004textrank} focus on extracting \textbf{present keyphrases} like ``relevance profiling'' in Figure~\ref{figure: keyphrase generation example}, which are the exact phrases appearing in the source text. The main ideas among them are \textit{identifying candidate phrases first and then ranking} algorithms. However, these methods ignore the semantic meaning underlying the context content and cannot generate \textbf{absent keyphrases} like ``interactive information retrieval'', which do not appear in the source text.

To overcome the above drawbacks, several encoder-decoder based keyphrase generation methods have been proposed including CopyRNN~\cite{meng2017dkg} and CopyCNN~\cite{zhang2017dkg_conv}. First, these methods treat the title and the main body equally and concatenate them as the only source text input. Then, the encoder maps each source text word into a hidden state vector which is regarded as the contextual representation. Finally, based on these representations, the decoder generates keyphrases from a predefined vocabulary regardless of the presence or absence of the keyphrases. 
A serious drawback of these models is that they ignore the leading role of the title and consequently fail to sufficiently utilize the already summarized information in it.

It is a widely agreed fact that the title can be viewed as a high-level summary of a document and the keyphrases provide more details of the key topics introduced in the document~\cite{li2010semi_title}. They play a similar and complementary role with each other. Therefore, keyphrases should have close semantic meaning to the title~\cite{li2010semi_title}. For example, as shown in Figure~\ref{figure: keyphrase generation example}, the title contains most of the salient points reflected by these keyphrases including ``retrieval'', ``profiling'', and ``evaluation''. Statistically, we study the proportion of keyphrases related to the title on the largest KG dataset and show the results in Table~\ref{table:statistic_study_on_val_set}. For simplicity, we define a \textit{TitleRelated} keyphrase as the keyphrase containing at least one common non-stop-word with the title. From Table~\ref{table:statistic_study_on_val_set}, we find that about 33\% absent keyphrases are \textit{TitleRelated}. For present keyphrases, the \textit{TitleRelated} percentage is up to around 60\%. By considering the fact that the length of a title is usually only 3\%-6\% of the corresponding source text, we can conclude that the title, indeed, contains highly summative and valuable information for generating keyphrases. 
Moreover, information in the title is also helpful in reflecting which part of the main body is essential, such as the part containing the same or related information with the title. For instance, in Figure~\ref{figure: keyphrase generation example}, the point ``evaluation'' in the title can guide us to focus on the part ``... task-oriented, comparative evaluation ...'' of the main body, which is highly related to the absent keyphrase ``task-oriented evaluation''.

To sufficiently leverage the title content, we introduce a new title-guided network by taking the above fact into the keyphrase generation scenario. In our model, the title is additionally treated as a query-like input in the encoding stage. First, two bi-directional Gated Recurrent Unit (GRU)~\cite{cho2014learning} layers are adopted to separately encode the context and the title into corresponding contextual representations. Then, an attention-based matching layer is used to gather the relevant title information for each context word according to the semantic relatedness. Since the context is the concatenation of the title and the main body, this layer implicitly contains two parts. The former part is the ``title to title'' self-matching, which aims to make the salient information in the title more important. The latter part is the ``main body to title'' matching wherein the title information is employed to reflect the importance of information in the main body. Next, an extra bi-directional GRU layer is used to merge the original contextual information and the corresponding gathered title information into the final title-guided representation for each context word. Finally, the decoder equipped with attention and copy mechanisms utilizes the final title-guided context representation to predict keyphrases. 

We evaluate our model on five real-world benchmarks, which test the ability of our model to predict present and absent keyphrases. Using these benchmarks, we demonstrate that our model can effectively exploit the title information and it outperforms the relevant baselines by a significant margin: for present (absent) keyphrase prediction, the improvement gain of F1-measure at 10 (Recall at 50) score is up to 9.4\% (19.1\%) compared to the best baseline on the largest dataset. Besides, we probe the performance of our model and a strong baseline CopyRNN on documents with different title length ratios (i.e., the title length over the context length). Experimental results show that our model consistently improves the performance with large gains, especially for documents with either very low or very high title length ratios.

\begin{table}[t]
\centering
\begin{tabular}{ c c c c}
\hline
\hline
 & Keyphrase & TitleRelated & \% \\
\hline
Present & 54,403 & 32,328 & 59.42 \\
\hline
Absent & 42,997  & 14,296 & 33.25\\
\hline
\end{tabular}
\caption{The statistics of \textit{TitleRelated} keyphrases on the validation set of \textbf{KP20k}.}
\label{table:statistic_study_on_val_set}
\end{table}

Our main contributions consist of three parts:
\begin{itemize}
\item A new perspective on keyphrase generation is explored, which sufficiently employs the title to guide the keyphrase prediction process.
\item A novel TG-Net model is proposed, which can effectively leverage the useful information in the title.
\item  The overall empirical results on five real-world benchmarks show that our model outperforms the state-of-the-art models significantly on both present and absent keyphrase prediction, especially for documents with either very low or very high title length ratios.
\end{itemize}

\section{Related Work} \label{section: related_work}
\subsection{Automatic Keyphrase Extraction}
Most of the automatic keyphrase extraction methods consist of two steps. Firstly, the \textit{candidate identification} step obtains a set of candidate phrases such as phrases with some specific part-of-speech (POS) tags~\cite{medelyan2009human,witten1999kea}. Secondly, in the \textit{ranking} step, all the candidates are ranked based on the importance computed by either unsupervised ranking approaches~\cite{wan2008single,mihalcea2004textrank,aaai_FlorescuC17_unsupervised} or supervised machine learning approaches~\cite{medelyan2009human,witten1999kea,nguyen2010wingnus_title,aaai_FlorescuJ18_supervised}. Finally, the top-ranked candidates are selected as the keyphrases. Besides these widely developed two-step approaches, there are also some methods using a sequence labeling operation to extract keyphrases~\cite{zhang2016keyphrase_twitter,luan2017scientific_seq_labeling,aaai_GollapalliLY17_seqlabel}. But they still cannot generate absent keyphrases.

Some extraction approaches~\cite{li2010semi_title,liu2011automatic_title} also consider the influence of the title. \citeauthor{li2010semi_title}~\shortcite{li2010semi_title} proposes a graph-based ranking algorithm which initializes the importance score of title phrases as one and the others as zero and then propagates the influence of title phrases iteratively. The biggest difference between \citeauthor{li2010semi_title}~\shortcite{li2010semi_title} and our method is that our method utilizes the contextual information of the title to guide the context encoding, while their model only considers the phrase occurrence in the title. \citeauthor{liu2011automatic_title}~\shortcite{liu2011automatic_title} models keyphrase extraction process as a translation operation from a document to keyphrases. The title is used as the target output to train the translator. Compared with our model, one difference is that this method still cannot handle semantic meaning of the context. The other is that our model regards the title as an extra query-like input instead of a target output.

\subsection{Automatic Keyphrase Generation}
Keyphrase generation is an extension of keyphrase extraction which explicitly considers the absent keyphrase prediction. CopyRNN~\cite{meng2017dkg} first frames the generation process as a sequence-to-sequence learning task and employs a widely used encoder-decoder framework~\cite{sutskever2014sequence2sequence} with attention~\cite{bahdanau2014attention} and copy~\cite{gu2016incorporating_copy} mechanisms. Based on CopyRNN, various extensions~\cite{HaiYe2018ssl,JunChen2018CorrRNN} are recently proposed. However, these recurrent neural network (RNN) based models may suffer the low-efficiency issues because of the computation dependency between the current time step and the preceding time steps in RNN. To overcome this shortcoming, CopyCNN~\cite{zhang2017dkg_conv} applies a convolutional neural network (CNN) based encoder-decoder model~\cite{gehring2017convolutional_seq2seq}. CopyCNN employs position embedding for obtaining a sense of order in the input sequence and adopts gated linear units (GLU)~\cite{dauphin2017language_glu} as the non-linearity function. CopyCNN not only achieves much faster keyphrase generation speed but also outperforms CopyRNN on five real-world benchmark datasets.

Nevertheless, both CopyRNN and CopyCNN treat the title and the main body equally, which ignores the semantic similarity between the title and the keyphrases. Motivated by the success of query-based encoding in various natural language processing tasks~\cite{gao2018generating,song2017unified,nema2017diversity,wang2017gated}, we regard the title as an extra query-like input to guide the source context encoding. Consequently, we propose a TG-Net model to explicitly explore the useful information in the title. In this paper, we focus on how to incorporate a title-guided encoding into the RNN-based model, but it is also convenient to apply this idea to the CNN-based model in a similar way.

\section{Problem Definition} \label{problem_definition}
We denote vectors with bold lowercase letters, matrices with bold uppercase letters and sets with calligraphy letters. We denote $\Theta$ as a set of parameters and $\mathbf{W}$ as a parameter matrix.

Keyphrase generation (KG) is usually formulated as follows: given a context $\mathbf{x}$, which is the concatenation of the title and the main body, output a set of keyphrases $\mathcal{Y}=\{\mathbf{y}^i\}_{i=1,\dots,M}$ where $M$ is the keyphrase number of $\mathbf{x}$.  
Here, the context $\mathbf{x}=[x_1,\dots,x_{L_{\mathbf{x}}}]$ and each keyphrase $\mathbf{y}^i=[y_1^i,\dots,y^i_{L_{\mathbf{y}^i}}]$ are both word sequences, where $L_{\mathbf{x}}$ is the length (i.e., the total word number) of the context and $L_{\mathbf{y}^i}$ is the length of the $i$-th produced keyphrase $\mathbf{y}^i$. 

To adapt the encoder-decoder framework, $M$ context-keyphrase pairs $\{(\mathbf{x}, \mathbf{y}^i)\}_{i=1,\dots,M}$ are usually split. Since we additionally use the title $\mathbf{t}=[t_1,\dots,t_{L_{\mathbf{t}}}]$ with length $L_{\mathbf{t}}$ as an extra query-like input, we split $M$ context-title-keyphrase triplets $\{(\mathbf{x}, \mathbf{t}, \mathbf{y}^i)\}_{i=1,\dots,M}$ instead of context-keyphrase pairs to feed our model. For conciseness, we use $(\mathbf{x}, \mathbf{t}, \mathbf{y})$ to represent such a triplet, where $\mathbf{y}$ is one of its target keyphrases.

\section{Our Proposed Model} \label{section: proposed_model}
\subsection{Title-Guided Encoder Module}
\begin{figure}[t]
\centering
\includegraphics[width=3.2in]{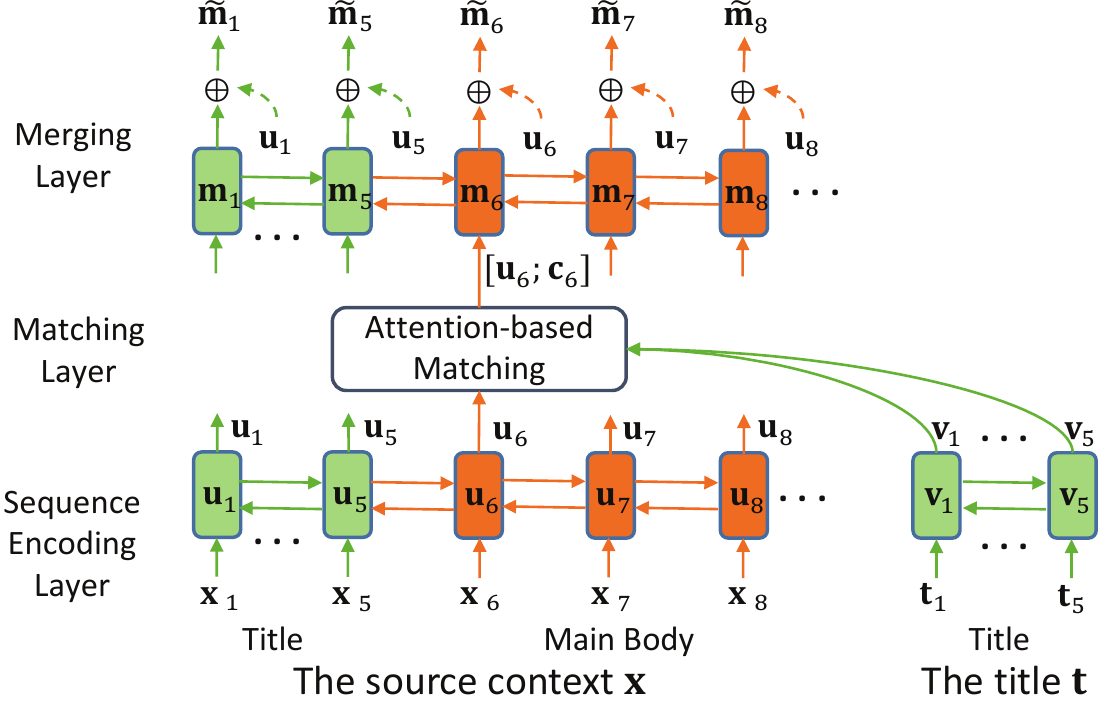}
\caption{The title-guided encoder module. (Best viewed in color.)}
\label{figure: title-guided encoder module}
\end{figure}

As shown in Figure~\ref{figure: title-guided encoder module}, the title-guided encoder module consists of a sequence encoding layer, a matching layer, and a merging layer. First, the sequence encoding layer reads the context input and the title input and learns their contextual representations separately. Then the matching layer gathers the relevant title information for each context word reflecting the important parts of the context. Finally, the merging layer merges the aggregated title information into each context word producing the final title-guided context representation.

\subsubsection{Sequence Encoding Layer}
At first, an embedding look-up table is applied to map each word within the context and the title into a dense vector with a fixed size $d_e$. To incorporate the contextual information into the representation of each word, two bi-directional GRUs~\cite{cho2014learning} are used to encode the context and the title respectively:
\begin{align}
\overrightarrow{\textbf{u}}_i &= \text{GRU}_{11}(\textbf{x}_i, \overrightarrow{\textbf{u}}_{i-1}), \\
\overleftarrow{\textbf{u}}_i  &= \text{GRU}_{12}(\textbf{x}_i, \overleftarrow{\textbf{u}}_{i+1}), \\
\overrightarrow{\textbf{v}}_j &= \text{GRU}_{21}(\textbf{t}_j, \overrightarrow{\textbf{v}}_{j-1}),\\
\overleftarrow{\textbf{v}}_j  &= \text{GRU}_{22}(\textbf{t}_j, \overleftarrow{\textbf{v}}_{j+1}),
\end{align}
where $i=1,2,\dots,L_{\mathbf{x}}$ and $j=1,2,\dots,L_{\mathbf{t}}$. $\mathbf{x}_i$ and $\mathbf{t}_j$ are the $d_e$-dimensional embedding vectors of the $i$-th context word and the $j$-th title word separately. $\overrightarrow{\textbf{u}}_i$, $\overleftarrow{\textbf{u}}_i$, $\overrightarrow{\textbf{v}}_j$, and $\overleftarrow{\textbf{v}}_j$ are $d/2$-dimensional hidden vectors where $d$ is the hidden dimension of the bi-directional GRUs. The concatenations $\mathbf{u}_i = [\overrightarrow{\mathbf{u}}_i; \overleftarrow{\mathbf{u}}_i] \in \mathbb{R}^d$ and $\mathbf{v}_j = [\overrightarrow{\mathbf{v}}_j; \overleftarrow{\mathbf{v}}_j] \in \mathbb{R}^d$ are used as the contextual vectors for the $i$-th context word and the $j$-th title word respectively.

\subsubsection{Matching Layer}
The attention-based matching layer is engaged to aggregate the relevant information from the title for each word within the context. The aggregation operation $\mathbf{c}_i = attn(\mathbf{u}_i, [\mathbf{v}_1, \mathbf{v}_2,\dots, \mathbf{v}_{L_{\mathbf{t}}}]; \mathbf{W}_1)$ is as follows:
\begin{align}
    \mathbf{c}_i &=  \sum_{j=1}^{L_{\mathbf{t}}} \alpha_{i,j} \mathbf{v}_j, \\
    \alpha_{i,j} &= \exp(s_{i,j}) \slash \sum_{k=1}^{L_{\mathbf{t}}} \exp(s_{i,k}),\\
    s_{i,j} &= (\mathbf{u}_i)^T \mathbf{W}_1 \mathbf{v}_j,
\end{align}
where $\mathbf{c}_i \in \mathbb{R}^d$ is the aggregated information vector for the $i$-th word of $\mathbf{x}$. $\alpha_{i,j}$ ($s_{i,j}$) is the normalized (unnormalized) attention score between $\mathbf{u}_i$ and $\mathbf{v}_j$.

Here, the matching layer is implicitly composed of two parts because the context is a concatenation of the title and the main body. The first part is the ``title to title'' self-matching part, wherein each title word attends the whole title itself and gathers the relevant title information. This part is used to strengthen the important information in the title itself, which is essential to capture the core information because the title already contains much highly summative information. The other part is the ``main body to title'' matching part wherein each main body word also aggregates the relevant title information based on semantic relatedness. In this part, the title information is employed to reflect the importance of information in the main body based on the fact that the highly title-related information in the main body should contain core information. 
Through these two parts, this matching layer can utilize the title information much more sufficiently than any of the previous sequence to sequence methods.

\subsubsection{Merging Layer}
Finally, the original contextual vector $\mathbf{u}_i$ and the aggregated information vector $\mathbf{c}_i$
are used as the inputs to another information merging layer:
\begin{align}
\overrightarrow{\mathbf{m}}_i &= \text{GRU}_{31}([\mathbf{u}_i; \mathbf{c}_i], \overrightarrow{\mathbf{m}}_{i-1}),\\
\overleftarrow{\mathbf{m}}_i &= \text{GRU}_{32}([\mathbf{u}_i; \mathbf{c}_i], \overleftarrow{\mathbf{m}}_{i+1}),\\
\widetilde{\mathbf{m}}_i &= \lambda \mathbf{u}_i + (1-\lambda) [\overrightarrow{\mathbf{m}}_i; \overleftarrow{\mathbf{m}}_i], \label{eq10}
\end{align}
where $[\mathbf{u}_i; \mathbf{c}_i] \in \mathbb{R}^{2d}$, $\overrightarrow{\mathbf{m}}_i \in \mathbb{R}^{d/2}$, 
$\overleftarrow{\mathbf{m}}_i \in \mathbb{R}^{d/2}$,
$[\overrightarrow{\mathbf{m}}_i, \overleftarrow{\mathbf{m}}_i] \in \mathbb{R}^d$,
and $\widetilde{\mathbf{m}}_i \in \mathbb{R}^d$. The $\mathbf{u}_i$ in Eq.~(\ref{eq10}) is a residual connection, and $\lambda \in (0,1)$ is the corresponding hyperparameter. Eventually, we obtain the title-guided contextual representation of the context (i.e., $[\widetilde{\mathbf{m}}_1, \widetilde{\mathbf{m}}_2,\dots, \widetilde{\mathbf{m}}_{L_{\mathbf{x}}}]$), which is regarded as a memory bank for the later decoding process. 

\subsection{Decoder Module}
After encoding the context into the title-guided representation, we engage an attention-based decoder~\cite{luong2015attention} incorporating with copy mechanism~\cite{see2017get_to_the_point} to produce keyphrases. Only one foward GRU is used in this module: 
\begin{align}
\mathbf{h}_t &= \text{GRU}_4([\mathbf{e}_{t-1}; \tilde{\mathbf{h}}_{t-1}], \mathbf{h}_{t-1}), \\
\hat{\mathbf{c}}_t &= attn(\mathbf{h}_t, [\widetilde{\mathbf{m}}_1, \widetilde{\mathbf{m}}_2,\dots, \widetilde{\mathbf{m}}_{L_{\mathbf{x}}}]; \mathbf{W}_2), \\
\tilde{\mathbf{h}}_{t} &= \text{tanh}(\mathbf{W}_3[\hat{\mathbf{c}}_t; \mathbf{h}_t]),
\end{align}
where $t=1,2,\dots,L_{\mathbf{y}}$, $\mathbf{e}_{t-1} \in \mathbb{R}^{d_e}$ is the embedding of the $(t-1)$-th predicted word wherein $\mathbf{e}_0$ is the embedding of the start token, $\hat{\mathbf{c}}_t \in \mathbb{R}^d$ is the aggregated vector for $\mathbf{h}_t \in \mathbb{R}^d$ from the memory bank $[\widetilde{\mathbf{m}}_1, \widetilde{\mathbf{m}}_2,\dots, \widetilde{\mathbf{m}}_{L_{\mathbf{x}}}]$, and $\tilde{\mathbf{h}}_t \in \mathbb{R}^d$ is the attentional vector at time step $t$.

Consequently, the predicted probability distribution over the predefined vocabulary $\mathcal{V}$ for current step is computed by:
\begin{align}
P_{v}(y_{t}|\mathbf{y}_{t-1}, \mathbf{x}, \mathbf{t}) = \text{softmax}(\mathbf{W}_v\tilde{\mathbf{h}}_{t} + \mathbf{b}_v),
\end{align}
where $\mathbf{y}_{t-1} = [y_1,\dots, y_{t-1}]$ is the previous predicted word sequence, and $\mathbf{b}_v \in \mathbb{R}^{|\mathcal{V}|}$ is a learnable parameter vector.

Before generating the predicted word, a copy mechanism is adopted to efficiently exploit the in-text information and to strengthen the extraction capability of our model. We follow \citeauthor{see2017get_to_the_point}~\shortcite{see2017get_to_the_point} and first calculate a soft switch between generating from the vocabulary and copying from the source context $\mathbf{x}$ at time step $t$:
\begin{equation}
g_t = \sigma(\mathbf{w}^T_g\tilde{\mathbf{h}}_{t} + b_g),
\end{equation}
where $\mathbf{w}_g \in \mathbb{R}^d$ is a learnable parameter vector and $b_g$ is a learnable parameter scalar.
Eventually, we get the final predicted probability distribution over the dynamic vocabulary $\mathcal{V}\cup \mathcal{X}$, where $\mathcal{X}$ are all words appearing in the source context. For simplicity, we use $P_v(y_t)$ and  $P_{final}(y_t)$ to denote $P_v(y_{t}|\mathbf{y}_{t-1}, \mathbf{x}, \mathbf{t})$ and $P_{final}(y_{t}|\mathbf{y}_{t-1}, \mathbf{x}, \mathbf{t})$ respectively:
\begin{equation}
P_{final}(y_t) = (1-g_t) P_v(y_t) + g_t\sum_{i:x_i=y_t}\hat{\alpha}_{t,i},
\end{equation}
where $\hat{\alpha}_{t,i}$ is the normalized attention score between $\mathbf{h}_t$ and $\widetilde{\mathbf{m}}_i$.
For all out-of-vocabulary (OOV) words (i.e., $y_t\notin \mathcal{V}$), we set $P_{v}(y_t)$ as zero. Similarly, if word $y_t$ does not appear in the source context $\mathbf{x}$ (i.e., $y_t\notin \mathcal{X}$), the copy probability $\sum_{i:x_i=y_t}\hat{\alpha}_{t,i}$ is set as zero.

\subsection{Training}
We use the negative log likelihood loss to train our model:
\begin{equation}
\mathcal{L} = -\sum_{t=1}^{L_{\mathbf{y}}} log P_{final}(y_t|\mathbf{y}_{t-1}, \mathbf{x}, \mathbf{t}; \Theta),
\end{equation}
where $L_{\mathbf{y}}$ is the length of target keyphrase $\mathbf{y}$ and $y_t$ is the $t$-th target word in $\mathbf{y}$, and $\Theta$ represents all the learnable parameters.

\section{Experiment Settings} \label{section: exp_settings}
The keyphrase prediction performance is first evaluated by comparing our model with the popular extractive methods and the state-of-the-art generative methods on five real-world benchmarks. Then, comparative experiments of different title length ratios are performed on our model and CopyRNN for further model exploration. Finally, an ablation study and a case study are conducted to better understand and interpret our model.

The experiment results lead to the following findings:
\begin{itemize}
\item Our model outperforms the state-of-the-art models on all the five benchmark datasets for both present and absent keyphrase prediction.
\item Our model consistently improves the performance on various title length ratios and obtains relative higher improvement gains for both very low and very high title length ratios.
\item The title-guided encoding part and the copy part are consistently effective in both present and absent keyphrase prediction tasks.
\end{itemize}
We implement the models using PyTorch~\cite{paszke2017pytorch} on the basis of the OpenNMT-py system~\cite{opennmt}.

\subsection{Training Dataset} \label{subsection:training_set}
Because of the public accessibility, many commonly-used scientific publication datasets are used to evaluate the explored KG methods. This study also focuses on generating keyphrases from scientific publications.
For all the generative models (i.e. our TG-Net model as well as all the encoder-decoder baselines), we choose the largest publicly available keyphrase generation dataset \textbf{KP20k} constructed by \citeauthor{meng2017dkg}~\shortcite{meng2017dkg} as the training dataset. \textbf{KP20k} consists of a large amount of high-quality scientific publications from various computer science domains. Totally 567,830 articles are collected in this dataset, where 527,830 for training, 20,000 for validation, and 20,000 for testing. Both the validation set and testing set are randomly selected. Since the other commonly-used datasets are too small to train a reliable generative model, we only train these generative models on \textbf{KP20k} and then test the trained models on all the testing part of the datasets listed in Table~\ref{table:statistic_testset}. As for the traditional supervised extractive baseline, we follow \citeauthor{meng2017dkg}~\shortcite{meng2017dkg} and use the dataset configuration shown in Table~\ref{table:statistic_testset}. To avoid the out-of-memory problem, for \textbf{KP20k}, we use the validation set to train the traditional supervised extractive baseline.

\subsection{Testing Datasets}
Besides \textbf{KP20k}, we also adopt other four widely-used scientific datasets for comprehensive testing, including \textbf{Inspec}~\cite{hulth2003improved}, \textbf{Krapivin}~\cite{krapivin2009large}, \textbf{NUS}~\cite{nguyen2007KE_sci_publications}, and \textbf{SemEval-2010}~\cite{kim2010semeval}. Table~\ref{table:statistic_testset} summarizes the statistics of each testing dataset.

\begin{table}[h]
\centering
\begin{tabular}{ c c c c}
\hline
\hline
 Dataset & Total & Training & Testing\\
\hline
\textbf{Inspec} & 2,000 & 1500 & 500\\
\hline
\textbf{Krapivin} & 2,304 & 1904 & 400\\
\hline
\textbf{NUS} & 211 & FFCV & 211\\
\hline
\textbf{SemEval-2010} & 288 & 188 & 100\\
\hline
\textbf{KP20k} & 567,830 & 20,000 & 20,000\\
\hline
\end{tabular}
\caption{The statistics of testing datasets. The ``Training'' means the training part for the traditional supervised extractive baseline. The ``FFCV'' represents five-fold cross validation. The ``Testing'' means the testing part for all models.}
\label{table:statistic_testset}
\end{table}

\subsection{Implementation Details}
For all datasets, the main body is the abstract, and the context is the concatenation of the title and the abstract. During preprocessing, various operations are performed including lowercasing, tokenizing by CoreNLP~\cite{manning2014stanford}, and replacing all the digits with the symbol $\langle digit \rangle$. We define the vocabulary $\mathcal{V}$ as the 50,000 most frequent words. 

We set the embedding dimension $d_e$ to 100, the hidden size $d$ to 256, and $\lambda$ to  0.5. All the initial states of GRU cells are set as zero vectors except that $\mathbf{h}_0$ is initialized as $[\overrightarrow{m}_{L_{\mathbf{x}}}; \overleftarrow{m}_1]$. We share the embedding matrix among the context words, the title words, and the target keyphrase words. All the trainable variables including the embedding matrix are initialized randomly with uniform distribution in [-0.1, 0.1]. The model is optimized by Adam~\cite{kingma2014adam} with batch size~=~64, initial learning rate~=~0.001, gradient clipping~=~1, and dropout rate~=~0.1. We decay the learning rate into the half when the evaluation perplexity stops dropping. Early stopping is applied when the validation perplexity stops dropping for three continuous evaluations. During testing, we set the maximum depth of beam search as 6 and the beam size as 200. We repeat the experiments of our model three times using different random seeds and report the averaged results.

We do not remove any predicted single-word phrase in the post-processing for \textbf{KP20k} during testing, which is different from \citeauthor{meng2017dkg}~\shortcite{meng2017dkg}, since our model is trained on this dataset and it can effectively learn the distribution of the single-word keyphrases. But for other testing datasets, we only keep the first predicted single-word phrase following \citeauthor{meng2017dkg}~\shortcite{meng2017dkg}.

\subsection{Baseline Models and Evaluation Metric}
For present keyphrase predicting experiment, we use two unsupervised models including TF-IDF and TextRank~\cite{mihalcea2004textrank}, and one supervised model Maui~\cite{medelyan2009human} as our traditional extraction baselines.
Besides, we also select CopyRNN~\cite{meng2017dkg} and CopyCNN~\cite{zhang2017dkg_conv}, the two state-of-the-art encoder-decoder models with copy mechanism~\cite{gu2016incorporating_copy}, as the baselines for present keyphrase prediction task. As for absent keyphrase prediction, since traditional extraction baselines cannot generate such keyphrases, we only choose CopyRNN and CopyCNN as the baseline models. For all baselines, we use the same setups as \citeauthor{meng2017dkg}~\shortcite{meng2017dkg} and  \citeauthor{zhang2017dkg_conv}~\shortcite{zhang2017dkg_conv}.

The \textit{recall} and \textit{F-measure} ($\text{F}_1$) are employed as our metrics for evaluating these algorithms. \textit{Recall} is the number of correctly predicted keyphrases over the total number of target kayphrases. $\text{F}_1$ score is computed based on the \textit{Recall} and  the \textit{Precision}, wherein \textit{Precision} is defined as the number of correctly predicted keyphrases over the total predicted keyphrase number. Following \citeauthor{meng2017dkg}~\shortcite{meng2017dkg} and \citeauthor{zhang2017dkg_conv}~\shortcite{zhang2017dkg_conv}, we also employ Porter Stemmer for preprocessing when determining whether two keyphrases are matched.

\begin{table*}
\centering
\begin{tabular}{>{\centering\arraybackslash}p{1.5cm}| 
>{\centering\arraybackslash}p{1.0cm} >{\centering\arraybackslash}p{1.1cm}| 
>{\centering\arraybackslash}p{1.0cm} >{\centering\arraybackslash}p{1.1cm}| 
>{\centering\arraybackslash}p{1.0cm} >{\centering\arraybackslash}p{1.1cm}| 
>{\centering\arraybackslash}p{1.0cm} >{\centering\arraybackslash}p{1.1cm}| 
>{\centering\arraybackslash}p{1.0cm} >{\centering\arraybackslash}p{1.1cm}}
 
  \hline
  \hline
 \multirow{2}{*}{\textbf{Model}} & \multicolumn{2}{c|}{\textbf{Inspec}} & \multicolumn{2}{c|}{\textbf{Krapivin}} & \multicolumn{2}{c|}{\textbf{NUS}} & \multicolumn{2}{c|}{\textbf{SemEval}} & \multicolumn{2}{c}{\textbf{KP20k}}
  \\
  & $\text{F}_1$@5 & $\text{F}_1$@10 & $\text{F}_1$@5 & $\text{F}_1$@10 & $\text{F}_1$@5 & $\text{F}_1$@10 & $\text{F}_1$@5 & $\text{F}_1$@10 & $\text{F}_1$@5 & $\text{F}_1$@10
  \\
  \hline
  \hline
  TF-IDF & 0.221 & 0.313 & 0.129 & 0.160 & 0.136 & 0.184 & 0.128 & 0.194 & 0.102 & 0.162
  \\
  TextRank & 0.223 & 0.281 & 0.189 & 0.162 & 0.195 & 0.196 & 0.176 & 0.187 & 0.175 & 0.147
  \\
  Maui & 0.040 & 0.042 & 0.249 & 0.216 & 0.249 & 0.268 & 0.044 & 0.039 & 0.270 & 0.230
  \\
  CopyRNN & 0.278 & 0.342 & 0.311 & 0.266 & 0.334 & 0.326 & 0.293 & 0.304 & 0.333 & 0.262
  \\
  CopyCNN & 0.285 & 0.346 & 0.314 & 0.272 & 0.342 & 0.330 & 0.295 & 0.308 & 0.351 & 0.288
  \\
  \hline
  TG-Net & \textbf{0.315} & \textbf{0.381} & \textbf{0.349} & \textbf{0.295} & \textbf{0.406} & \textbf{0.370} & \textbf{0.318} & \textbf{0.322} & \textbf{0.372} & \textbf{0.315}
  \\
  \% gain & 10.5\% & 10.1\% & 11.1\% & 8.5\% & 18.7\% & 12.1\% & 7.8\% & 4.5\% & 6.0\% & 9.4\% 
 \\
 \hline
\end{tabular}
\caption{Present keyphrase predicting results on all test datasets. ``\% gain'' is the improvement gain over CopyCNN.}
\label{Table:present_keyphrase_results}
\end{table*}

\begin{table*} 
\centering
\begin{tabular}{>{\centering\arraybackslash}p{1.5cm}| 
>{\centering\arraybackslash}p{1.0cm} >{\centering\arraybackslash}p{1.1cm}| 
>{\centering\arraybackslash}p{1.0cm} >{\centering\arraybackslash}p{1.1cm}| 
>{\centering\arraybackslash}p{1.0cm} >{\centering\arraybackslash}p{1.1cm}| 
>{\centering\arraybackslash}p{1.0cm} >{\centering\arraybackslash}p{1.1cm}| 
>{\centering\arraybackslash}p{1.0cm} >{\centering\arraybackslash}p{1.1cm}}
  \hline
  \hline
  \multirow{2}{*}{\textbf{Model}} & \multicolumn{2}{c|}{\textbf{Inspec}} & \multicolumn{2}{c|}{\textbf{Krapivin}} & \multicolumn{2}{c|}{\textbf{NUS}} & \multicolumn{2}{c|}{\textbf{SemEval}} & \multicolumn{2}{c}{\textbf{KP20k}}
  \\
  & R@10 & R@50 & R@10 & R@50 & R@10 & R@50 & R@10 & R@50 & R@10 & R@50
  \\
  \hline
  \hline
  CopyRNN & 0.047 & 0.100 & 0.113 & 0.202 & 0.058 & 0.116 & 0.043 & 0.067 & 0.125 & 0.211
  \\
  CopyCNN & 0.050 & 0.107 & 0.119 & 0.205 & 0.062 & 0.120 & 0.044 & 0.074 & 0.147 & 0.225
  \\
  \hline
  TG-Net & \textbf{0.063} & \textbf{0.115}  & \textbf{0.146} & \textbf{0.253} & \textbf{0.075} & \textbf{0.137} & \textbf{0.045} & \textbf{0.076} & \textbf{0.156} & \textbf{0.268}
  \\
  \% gain & 26.0\% & 7.5\% & 22.7\% & 23.4\% & 21.0\% & 14.2\%
 & 2.3\% & 2.7\% & 6.1\% & 19.1\% 
 \\
  \hline
\end{tabular}
\caption{Absent keyphrase predicting results on all test datasets. ``\% gain'' is the improvement gain over CopyCNN.}
\label{Table:absent_keyphrase_prediction}
\end{table*}

\section{Results and Analysis} \label{section: Results and Analysis}
\subsection{Present Keyphrase Predicting}
In this section, we compare present keyphrase prediction ability of these models on the five real-world benchmark datasets. The F-measures at top 5 and top 10 predictions of each model are shown in Table~\ref{Table:present_keyphrase_results}.

From this table, we find that all the generative models significantly outperforms all the traditional extraction baselines. Besides, we also note that our TG-Net model achieves the best performance on all the datasets with significant margins. For example, on \textbf{KP20k} dataset, our model improves 9.4\% ($\text{F}_1$@10 score) than the best generative model CopyCNN. Compared to CopyRNN which also applies an RNN-based framework, our model improves about 20.2\%. The results show that our model obtains much stronger keyphrase extraction ability than CopyRNN and CopyCNN.

\subsection{Absent Keyphrase Predicting}
In this setting, we consider the absent keyphrase predicting ability which requires the understanding of the semantic meaning of the context. Only the absent target keyphrases and the absent predictions are preserved for this evaluation. Generally, recalls at top 10 and top 50 predictions are engaged as the metrics to evaluate how many absent target keyphrases are correctly predicted. 

The performance of all models is listed in Table~\ref{Table:absent_keyphrase_prediction}. It is observed that our TG-Net model consistently outperforms the previous sequence-to-sequence models on all the datasets. For instance, our model exceeds 19.1\% (R@50 score) on \textbf{KP20k} than the state-of-the-art model CopyCNN. Overall, the results indicate that our model is able to capture the underlying semantic meaning of the context content much better than these baselines, as we have anticipated.

\begin{figure}[t]
\centering
\includegraphics[width=3.2in]{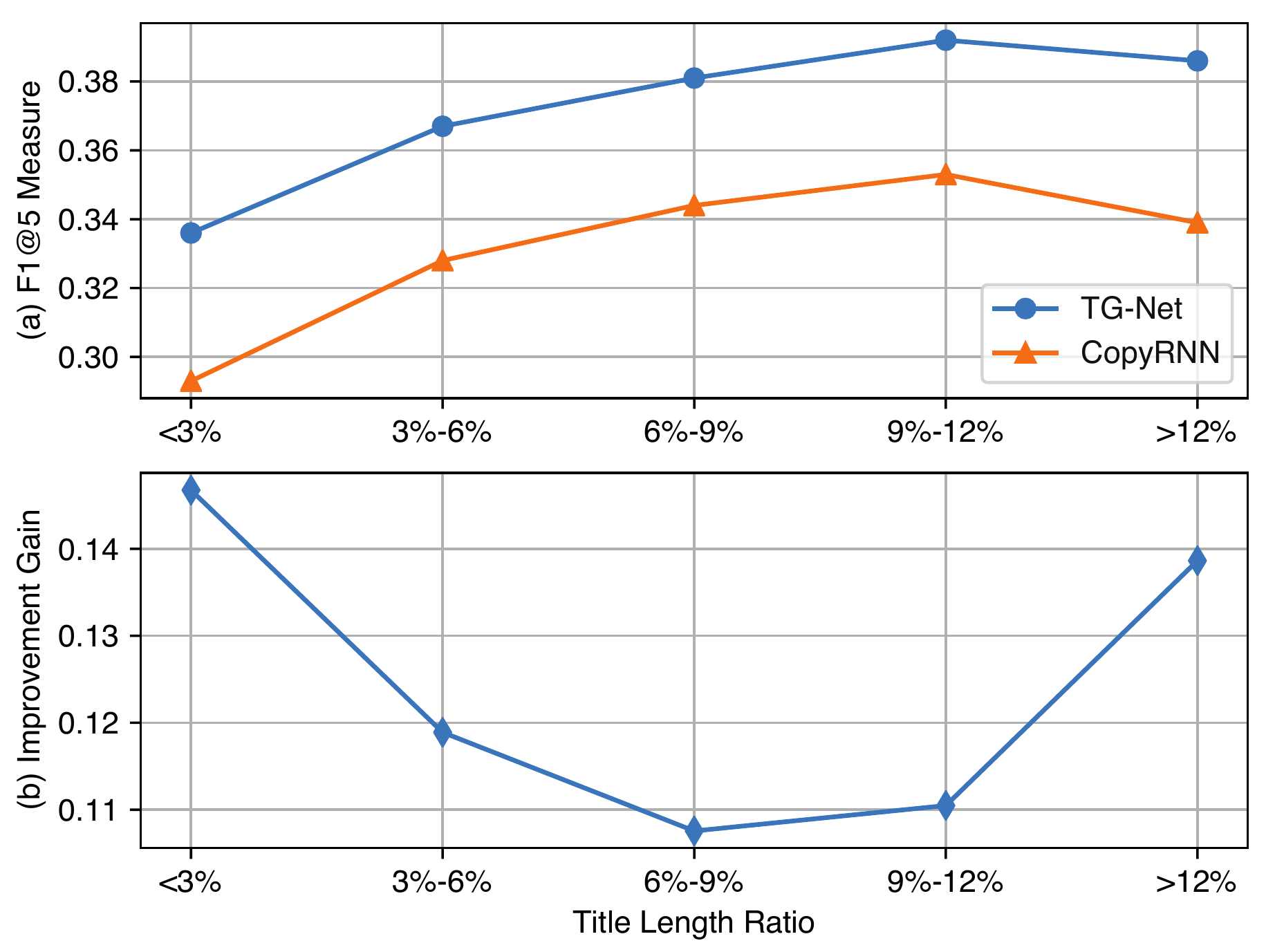}
\caption{Present keyphrase predicting ability (F1@5 measure) on various title length ratios.}
\label{figure:title_lens_present_cureve}
\end{figure}

\begin{figure*}
\centering
\includegraphics[width=6.9in]{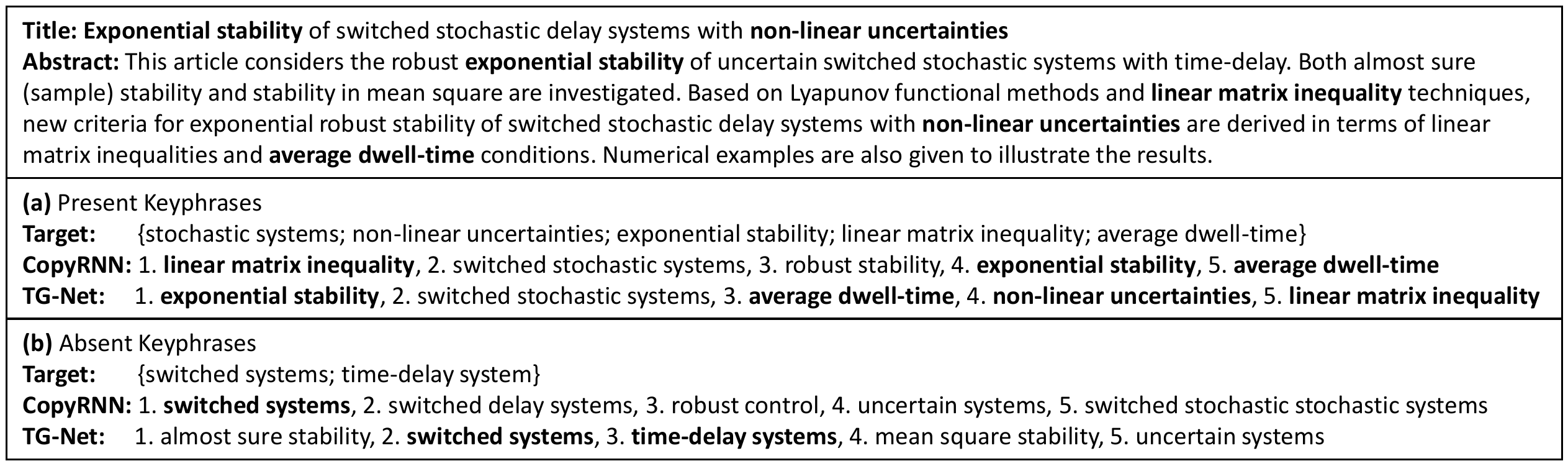}
\caption{A prediction example of CopyRNN and TG-Net. The top 5 predictions are compared and the correct predictions are highlighted in bold.}
\label{figure:case_study}
\end{figure*}

\subsection{Keyphrase Predicting on Various Title Length Ratios}
To find out how our title incorporation influences the prediction ability, we compare the keyphrase predicting ability of two RNN-based models (i.e., our model and CopyRNN) on different title length ratios. The title length ratio is defined as the title length over the context length. This analysis is based on the \textbf{KP20k} testing dataset. In view of the title length ratio, we preprocess the testing set into five groups (i.e., $<$3\%, 3\%-6\%, 6\%-9\%, 9\%-12\% and $>$12\%). Then, the present keyphrase prediction results (F1@5 measure) and the improvement gain on each group are depicted in Figure~\ref{figure:title_lens_present_cureve}.

In Figure~\ref{figure:title_lens_present_cureve}(a), we notice that both CopyRNN and our TG-Net model generally perform better when the title length ratio is higher. One possible explanation is that when the title is long, it conveys substantial salient information of the abstract. Therefore, the chance for the models to attend to the core information is enhanced, which leads to the observed situation. This figure also shows that both TG-Net and CopyRNN get worse performance on $>$12\% group than 9\%-12\% group. The main reason is that there exist some data with a short abstract in $>$12\% group, which leads to the lack of enough context information for correctly generating all keyphrases.

In Figure~\ref{figure:title_lens_present_cureve}(b), we find that our TG-Net consistently improves the performance with a large margin on five testing groups, which again indicates the effectiveness of our model. In a finer perspective, we note that the improvement gain is higher on the lowest (i.e., $<$3\%) and the highest (i.e., $>$12\%) title length ratio groups. In $>$12\% group, the title plays a more important role than in other groups, and consequently our model benefits more by not only explicitly emphasizing the title information itself, but also utilizing it to guide the encoding of information in the main body. As for $<$3\% group, the effect of such a short title is small on the latter part of the context in CopyRNN because of the long distance. However, our model explicitly employs the title to guide the encoding of each context word regardless of the distance, which utilizes the title information much more sufficiently. Consequently, our model achieves much higher improvement in this group. While we only display the results of present keyphrase prediction, the absent keyphrase predicting task gets the similar results. 

\subsection{Ablation Study}
We also perform an ablation study on \textbf{Krapivin} for better understanding the contributions of the main parts of our model. For a comprehensive comparison, we conduct this study on both present keyphrase prediction and absent keyphrase prediction. 

As shown in Table~\ref{table:ablation study}, after we remove the title-guided part and only reserve the sequence encoding for the context (i.e., -title), both the present and absent keyphrase prediction performance become obviously worse, indicating that our title-guided context encoding is consistently critical for both present and absent keyphrase generation tasks. We also investigate the effect of removing the copy mechanism (i.e., -copy) from our TG-Net. From Table~\ref{table:ablation study}, we notice that the scores decrease dramatically on both present and absent keyphrase prediction, which demonstrates the effectiveness of the copy mechanism in finding important parts of the context.

\subsection{Case Study}
A keyphrase prediction example for a paper about the exponential stability of uncertain switched stochastic delay systems is shown in Figure~\ref{figure:case_study}. To be fair, we also only compare the RNN-based models (i.e., TG-Net and CopyRNN). For present keyphrase, we find that a present keyphrase ``non-linear uncertainties'', which is a title phrase, is correctly predicted by our TG-Net, while CopyRNN fails to do so. As for absent keyphrase, we note that CopyRNN fails to predict the absent keyphrase ``time-delay systems''. But our TG-Net can effectively utilize the title information ``stochastic delay systems'' to locate the important abstract information ``stochastic systems with time-delay'' and then successfully generate this absent keyphrase. These results exhibit that our model is capable of capturing the title-related core information more effectively and achieving better results in predicting present and absent keyphrases.

\begin{table}[t]
\centering
\begin{tabular}{c| c c| c c}
  \hline
  \hline
  & \multicolumn{2}{c|}{\textbf{Present}} & \multicolumn{2}{c}{\textbf{Absent}}
  \\
  \textbf{Model} & F1@5 & F1@10 & R@10 & R@50
  \\
  \hline
  \hline
  TG-Net & \textbf{0.349} & \textbf{0.295} & \textbf{0.146} & \textbf{0.253}\\
  \hline
 -title & 0.334 & 0.288 & 0.142 & 0.240\\
  \hline
 -copy & 0.306 & 0.281 & 0.127 & 0.216\\
  \hline
\end{tabular}
\caption{Ablation study on \textbf{Krapivin} dataset.}
\label{table:ablation study}
\end{table}

\section{Conclusion} \label{section: conclusion}
In this paper, we propose a novel TG-Net for keyphrase generation task, which explicitly considers the leading role of the title to the overall document main body. Instead of simply concatenating the title and the main body as the only source input, our model explicitly treats the title as an extra query-like input to guide the encoding of the context. The proposed TG-Net is able to sufficiently leverage the highly summative information in the title to guide keyphrase generation. The empirical experiment results on five popular real-world datasets exhibit the effectiveness of our model for both present and absent keyphrase generation, especially for a document with very low or very high title length ratio. One interesting future direction is to explore more appropriate evaluation metrics for the predicted keyphrases instead of only considering the exact match with the human labeled keyphrases as the current \textit{recall} and \textit{F-measure} do.

\section{Acknowledgments}
The work described in this paper was partially supported
by the Research Grants Council of the Hong
Kong Special Administrative Region, China (No. CUHK
14208815 and No. CUHK 14210717 of the General Research
Fund), and Microsoft Research Asia (2018 Microsoft
Research Asia Collaborative Research Award). We would like to thank Jingjing Li, Hou Pong Chan, Piji Li and Lidong Bing for their comments.

\fontsize{9.0pt}{10.0pt} \selectfont
\bibliography{PaperID3210_TGNet}
\bibliographystyle{aaai}

\end{document}